%% file: main.tex
\documentclass[10pt,twocolumn,letterpaper]{article}

\usepackage{cvpr}
\usepackage{times}
\usepackage{epsfig}
\usepackage{graphicx}
\usepackage{amsmath}
\usepackage{amssymb}
\usepackage[utf8]{inputenc}

\usepackage{pdflscape}
\usepackage{multirow}
\usepackage[table,xcdraw]{xcolor}

\usepackage{flushend} 
\usepackage[numbers, sort]{natbib}

\usepackage[symbol]{footmisc}

\usepackage{perpage} 
\MakePerPage{footnote} 

\usepackage[breaklinks=true,bookmarks=false]{hyperref}

\cvprfinalcopy 

\def\cvprPaperID{****} 

\begin{document}

\title{Method Towards CVPR 2021 Image Matching Challenge}

\author{Xiaopeng Bi\footnotemark[2],  Ran Yan, Zheng Chai, Haotian Zhang, Xiao Liu\footnotemark[4]\\
Megvii Inc. Research 3D\\
{\tt\small \{bixiaopeng, yanran, chaizheng, zhanghaotian, liuxiao\}@megvii.com}

}

\maketitle

\begin{abstract}
This report describes Megvii-3D team’s approach towards SimLocMatch Challenge @ CVPR 2021 Image Matching Workshop. It includes the submissions:

\begin{itemize}
\setlength\itemsep{-0.5em}
\item \href{https://simlocmatch.com/show_public_submission/c65b43a8-d1dd-11eb-a9db-9b02b04a8ee2}{aaa-1000k\_50\_no\_ms111}
\item \href{https://simlocmatch.com/show_public_submission/16b4f74c-d1e1-11eb-a9db-9b02b04a8ee2}{aaa-1000k\_80\_no\_ms111}
\item \href{https://simlocmatch.com/show_public_submission/acc5eeee-d1c8-11eb-a9db-9b02b04a8ee2}{aaa-1000k\_no\_ms}
\item \href{https://simlocmatch.com/show_public_submission/abbe9ac2-d1d3-11eb-a9db-9b02b04a8ee2}{aaa-1000k\_no\_ms2}
\item \href{https://simlocmatch.com/show_public_submission/efd3fe64-d192-11eb-a9db-9b02b04a8ee2}{sss-sd\_100k\_8}
\item \href{https://simlocmatch.com/show_public_submission/2e7a7c8a-d047-11eb-a9db-9b02b04a8ee2}{sss-sd\_100k\_1}
\item \href{https://simlocmatch.com/show_public_submission/3c207798-d1c5-11eb-a9db-9b02b04a8ee2}{sss-sd\_100k\_6}
\end{itemize}

The methodology we took is similar to our solution towards CVPR 2021 Image Matching Challenge, hence we only report the additional strategies and tricks besides it.

\end{abstract}

\footnotetext{
\footnotemark[2] Work conducted during their internships at Megvii 3D
}
\footnotetext{
\footnotemark[4] Corresponding author
}

\section*{Method}
For the simloc matching Challenge, we tried DISK\cite{tyszkiewicz2020disk} 8k keypoints with its SuperGlue\cite{sarlin20superglue} matcher, as well as the Superpoint\cite{superpoint_paper} 2k keypoints together with DISK 6k keypoints as a combination, where each feature was matched by its corresponding SugerGlue matching. We noticed that the latter outperformed the former by a obvious margin, so we stuck with the second approach.

We compared the single-scale result with the multiple-scale one and found that the first one has better performance in the number of inliers and matching success ratio, whereas the last one worked better in terms of the number of non-matches. This founding agreed with our intuition, as the more the number of matching is, the more the number of inliers potentially, which leads to a later superior matching success ratio. Also, mixing multiple-scale matching outcomes together would result in a boost of the number of matches for non-matching images.

Since the simloc dataset contains two indoor scenes, office and restaurant, which fits the indoor SuperGlue weights better. Due to the limitation of time, we did not have enough resources to train a DISK-SuperGlue matcher for the indoor scenes, thus we only applied the indoor weights of SuperPoint-SuperGlue provided by the original author. 

In order to further reduce the number of matches in non-matching images, we decided to set a threshold regarding the number of matches between each pair, as the 'discarding threshold'. We would indicate that 8 is a theoretical number as if the number of matches is less than 8, the pose cannot be solved. While other thresholds are practical numbers. We carried out an analysis of the distribution of the number of matches for each pair and conducted some experiments. 

The outlier rejection was implemented by DegenSAC\cite{Chum2005} without further tuning.

{\small
\bibliographystyle{IEEEtran}
\bibliography{egbib}
}

\newpage
\begin{landscape}

\section*{Appendix: Details about each Submission}

\input{table}

\end{landscape}

\end{document}

%% file: table.tex
\begin{table}[ht!]
\resizebox{\columnwidth}{!}{%
\begin{tabular}{|l|l|l|l|l|l|l|l|l|l|l|l|l|l|l|l|}
\hline
\multicolumn{1}{|c|}{\textbf{methods}} &
  \multicolumn{1}{c|}{\textbf{image size(sp disk)}} &
  \multicolumn{1}{c|}{\textbf{scannet}} &
  \multicolumn{1}{c|}{\textbf{discard nums}} &
  \multicolumn{1}{c|}{\textbf{degensac th}} &
  \multicolumn{1}{c|}{\textbf{scale}} &
  \multicolumn{1}{c|}{\textbf{disk nms}} &
  \multicolumn{1}{c|}{\textbf{sp nms}} &
  \multicolumn{1}{c|}{\textbf{disk max keypoints}} &
  \multicolumn{1}{c|}{\textbf{sp max keypoints}} &
  \multicolumn{1}{c|}{\textbf{disk match score}} &
  \multicolumn{1}{c|}{\textbf{sp  match score}} &
  \multicolumn{1}{c|}{\textbf{degensac iter}} &
  \multicolumn{1}{c|}{\textbf{inlieres}} &
  \multicolumn{1}{c|}{\textbf{match success rate}} &
  \multicolumn{1}{c|}{\textbf{matches(non-matches)}} \\
\hline
{\color[HTML]{0000EE} sss-sd\_100k\_1}          & 1600/1600 & N       & 0  & 1.1 & Y/Y & 4 & 4 & 6000 & 2048 & 0.7 & 0.2 & 100k  & 248.66 & 51.02\% & 47.72 \\
\hline
{\color[HTML]{0000EE} sss-sd\_100k\_6}          & 1600/1600 &  Y & 0  & 1.1 & Y/Y & 4 & 4 & 6000 & 2048 & 0.7 & 0.2 & 1000k & 241.23 & 50.12\% & 36.99 \\
\hline
{\color[HTML]{0000EE} sss-sd\_100k\_8}          & 1600/1600 & N       & 0  & 1.1 & Y/Y & 4 & 4 & 6000 & 4096 & 0.7 & 0.2 & 1000k & 319.02 & 52.19\% & 47.23 \\
\hline
{\color[HTML]{0000EE} aaa-1000k\_no\_ms}        & 1600/1600 & N       & 8  & 1.1 & N/N & 4 & 4 & 6000 & 4096 & 0.7 & 0.2 & 1000k & 314.43 & 51.04\% & 30.01 \\
\hline
{\color[HTML]{0000EE} { aaa-1000k\_no\_ms2}} & 1600/1600 & N       & 50 & 1.1 & N/N & 4 & 4 & 6000 & 4096 & 0.7 & 0.2 & 1000k & 312.75 & 44.63\% & 23.15 \\
\hline
{\color[HTML]{0000EE} aaa-1000k\_80\_no\_ms111} & 1600/1600 & N       & 8  & 0.8 & N/N & 4 & 4 & 6000 & 4096 & 0.7 & 0.2 & 100k  & 275.39 & 51.40\% & 26.30 \\
\hline
{\color[HTML]{0000EE} aaa-1000k\_50\_no\_ms111} & 1600/1600 & N       & 8  & 0.5 & N/N & 4 & 4 & 6000 & 4096 & 0.7 & 0.2 & 100k  & 214.70 & 51.56\% & 21.80
\\ \hline
\end{tabular}}

\caption{Submission Details}
\end{table}